\documentclass[letterpaper, 10 pt, conference]{ieeeconf}  

\usepackage{amsmath}
\usepackage{amssymb}
\usepackage{url}
\usepackage{graphicx}

\usepackage{color}
 
\usepackage{caption}
\usepackage{subcaption}
\usepackage{booktabs}
\usepackage{threeparttable}

\usepackage[noadjust]{cite}

\usepackage{bbding}

\let\labelindent\relax 
\usepackage[inline]{enumitem}

\IEEEoverridecommandlockouts                              

\title{\LARGE \bf
ECTLO: Effective Continuous-time Odometry Using Range Image for LiDAR with Small FoV
}

\author{Xin~Zheng and
        Jianke~Zhu
\thanks{Xin~Zheng and Jianke~Zhu are with the College of Computer Science, Zhejiang University, Hangzhou, China, 310027. \protect\\ 
E-mail: \{xinzheng,jkzhu\}@zju.edu.cn.}
\thanks{Jianke Zhu is the Corresponding Author.}}

\begin{document}

\maketitle
\thispagestyle{empty}
\pagestyle{empty}

\begin{abstract}

Prism-based LiDARs are more compact and cheaper than the conventional mechanical multi-line spinning LiDARs, which have become increasingly popular in robotics, recently. However, there are several challenges for these new LiDAR sensors, including small field of view, severe motion distortions, and irregular patterns, which hinder them from being widely used in LiDAR odometry, practically. To tackle these problems, we present an effective continuous-time LiDAR odometry (ECTLO) method for the Risley-prism-based LiDARs with non-repetitive scanning patterns. A single range image covering historical points in LiDAR's small FoV is adopted for efficient map representation. To account for the noisy data from occlusions after map updating, a filter-based point-to-plane Gaussian Mixture Model is used for robust registration. Moreover, a LiDAR-only continuous-time motion model is employed to relieve the inevitable distortions. Extensive experiments have been conducted on various testbeds using the prism-based LiDARs with different scanning patterns, whose promising results demonstrate the efficacy of our proposed approach.

\end{abstract}

\section{INTRODUCTION}

Light detection and ranging (LiDAR) sensors can directly obtain the accurate range measurements in various scenarios by actively emitting the laser beams, which enables them to be the essential sensors for perception and navigation. Currently, the dominant LiDAR odometry approaches~\cite{zhang2014loam,shan2018lego,deschaud2018imls,behley2018efficient,zheng2021efficient} make use of the multi-line mechanical spinning scanners due to their simplicity and success in many robotic applications~\cite{thrun2006stanley,reinke2022iros}.

With the prevalence of autonomous driving, there is a demand for consumer-grade vehicle-borne LiDAR. However, the conventional multi-line spinning LiDAR cannot fulfill the massive deployment requirements on price, size and reliability. To this end, micro-electro-mechanical systems (MEMS) and rotating prism are two major alternative techniques used in the consumer market. Both of them try to reduce their sizes by using fewer pairs of semiconductor transceivers, and a mechanism to change the light direction at high frequency in order to cover the area in the limited field of view (FoV), which makes them appropriately integrate into the car. In this paper, we focus our attention on the Risley-prism-based LiDARs~\cite{liu2021low}. Despite that recent works~\cite{lin2020loam,li2021towards,xu2021fast,lin2021r3live} have adopted this LiDAR into odometry task, they are still variant methods of multi-line spinning LiDAR, and cannot take full consideration of the new characteristics of prism-based LiDARs. 

\begin{figure}[t]
	\centering
	\includegraphics[width=0.5\textwidth]{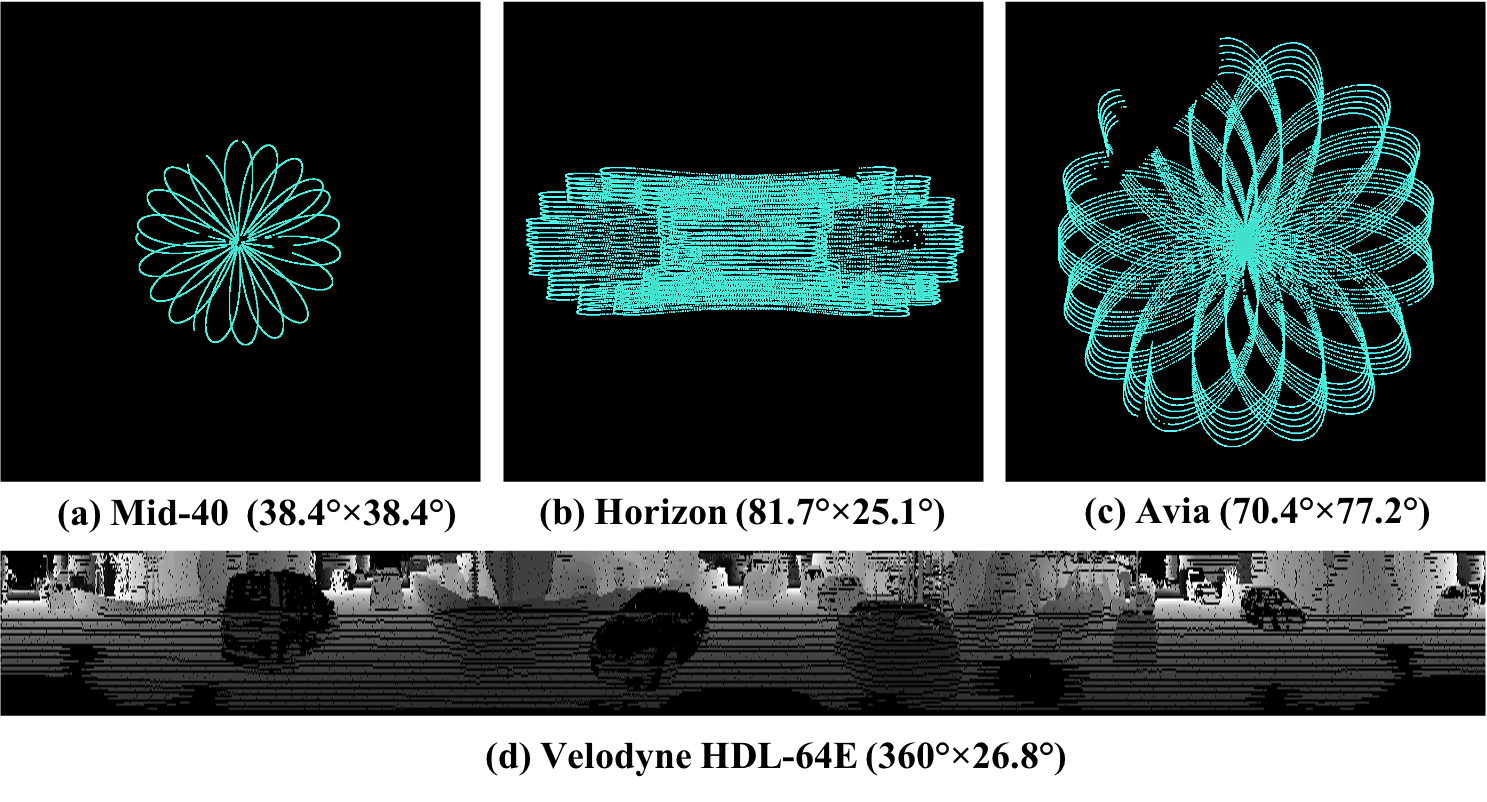}
	\caption{\footnotesize Spherical projection results of different LiDARs within 0.1 second of integration duration. Compared to spinning LiDAR (d), the scanning patterns of SSL (a-c) are sparse with a small FoV.  }
	\label{fig:pattern}
	\vspace{-0.2in}
\end{figure}

A significant property of prism-based LiDAR is its limited horizontal FoV, as shown in Fig.~\ref{fig:pattern}, which was previously regarded as a drawback for registration. In this paper, we find this property can be utilized for a compact map structure. Current LiDAR odometry methods~\cite{lin2020loam,vizzo2023ral} usually maintain a local map and adopt a scan-to-map strategy for more precise registration. Conventional data structures for map representation are either KD-tree~\cite{lin2020loam,xu2021fast} or voxel hash table~\cite{dellenbach2022ct,bai2022faster,vizzo2023ral}, which may not be the optimal choice for LiDARs with small FoV. The fundamental problem of LiDAR odometry is the registration between two consecutive point clouds, so only the overlap region affects the final optimization. Keeping all historical points around the current center within a fixed radius is unnecessary. The region of interest lies in an area even smaller than LiDAR's FoV. Thus, we extend a 2D  range image map representation for prism-based LiDAR, which has been proven accurate and efficient for spinning LiDAR in diverse scenarios~\cite {zheng2021efficient,cowley2021upslam,qu2022llol}. The image map is robocentric, and only covers the historical points in LiDAR's FoV. This map structure is memory efficient and friendly for the parallel implementation. 

Unfortunately, directly bringing the idea of image map representation is ineffective for LiDARs with  non-repetitive scanning pattern. Since occlusions are inevitable in this 2D map structure, the map will get quite noisy with continuous updating. Another common issue for different kinds of LiDAR is motion distortions, which is more serious in this new sensor affected by non-repetitive scanning mode. Moreover, the diverse sparse scanning patterns make the feature extraction complicated as in~\cite{zhang2014loam,shan2018lego}.

\begin{figure}[t]
	\centering
	\includegraphics[width=0.5\textwidth]{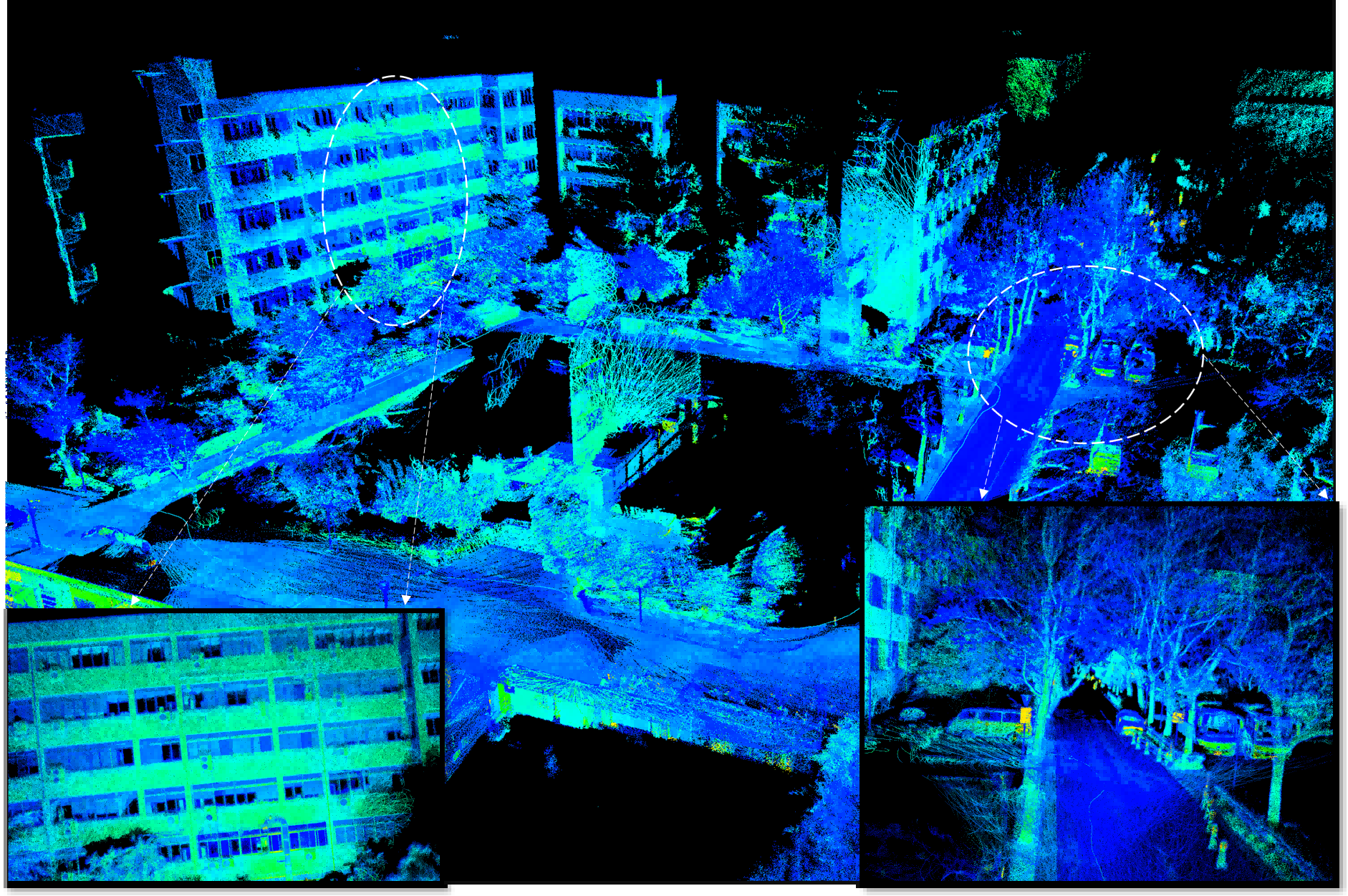}
	\caption{\footnotesize Reconstruction results of our proposed approach on the data collected by Livox Mid-40. Two pictures at the bottom show the details of structured facades and unstructured streets. } 
	\label{fig:map}
	\vspace{-0.2in}
\end{figure}

To tackle the above critical issues, we present an effective Continuous Time LiDAR Odometry (ECTLO) method for LiDAR with small FoV in this paper. Without extra sensors, a continuous-time motion model~\cite{dellenbach2022ct} is applied during pose optimization. To account for the noisy data after 2D image map updating, we employ a point-to-plane Gaussian Mixture Model scheme~\cite{gao2019filterreg} to effectively align the scans. In this framework, we directly register raw points from LiDAR to avoid the complicated feature extraction for the diverse scanning patterns. Fig.~\ref{fig:map} shows the mapping result using a single Livox Mid-40 with a $38.4^{\circ}$ sparse circular pattern.

In summary, the main contributions of this paper are: 1) an effective odometry method for LiDAR with small FoV by taking advantage of range image map representation; 2) a novel robust continuous-time filter registration scheme handling motion distortions and diverse sparse scanning patterns; 3) extensive experiments on a series of challenging datasets demonstrate that our proposed approach outperforms the state-of-the-art LiDAR-only odometry methods. 

\section{RELATED WORKS}\label{sec:rel}
Our proposed approach aims to deal with the challenge of LiDAR with small FoV in odometry task. In this section, we briefly review the most relevant studies on prism-based LiDAR odometry, motion compensation, and point set registration.
\subsection{Prism-based LiDAR Odometry}
Depending on the prism combinations, the prism LiDARs have different patterns that are irregular and non-repetitive, as shown in Fig.~\ref{fig:pattern}. Comparing to the conventional multi-line LiDARs with 360-degree coverage, these LiDARs have the sparse scanning patterns with small FoVs, which raises many issues on scan registration. Lin and Zhang~\cite{lin2020loam} firstly employ a prism-based LiDAR Livox Mid-40 for outdoor scanning. Liu and Zhang~\cite{liu2021balm} adopt a bundle adjustment framework with Livox Horizon for structural scene.

Due to the small FoV and low density of point cloud, the conventional methods cannot handle the fast motion and jittering. To overcome the limitation of single sensor, Li et al.~\cite{li2021towards} present a tightly-coupled LiDAR-inertial odometry framework for Livox Horizon. Moreover, Xu and Zhang~\cite{xu2021fast} fuse multi-sensor information by an Iterated Extended Kalman Filter using Livox Avia. To achieve the accurate registration, both of them rely on the additional sensor and LiDARs having more dense scanning patterns with large FoV.

Those methods maintain a local map in KD-Tree that store points near the current LiDAR center within a fixed radius. However, overlap regions between the consecutive scans are restricted in the sensor's FoV. It is unnecessary to keep such abundant historical points for LiDAR with small FoV. Thus, we use a 2D robocentric image in this paper for efficient representation~\cite{zheng2021efficient}.

\subsection{Motion Compensation}
Motion distortions in LiDARs are mainly due to the continuous movement of laser beam. To address this issue, a straightforward approach is to make use of the constant velocity motion model~\cite{zhang2014loam, deschaud2018imls}, which undistorts the current point cloud with velocity from previous poses or extra IMU measurements. However, the constant velocity assumption may not be valid in real-world applications with fast orientation or position changes, especially for the handheld devices. Continuous-time trajectory~\cite{park2018elastic,droeschel2018efficient,furgale2012continuous} is more accurate to represent the nature continuous motions. To account for fast motion, it requires a lot of control poses within a scan. Meanwhile, the strict continuity of spline between control poses may be sensitive to noise during optimization. These drawbacks prohibit it from the real-time applications. Recently, Dellenbach et al.~\cite{dellenbach2022ct} compensate the distortions in the conventional spinning LiDAR by interpolating the beginning pose with the end one within a scan, where the discontinuity between consecutive scans is penalized by the additional position and velocity constraints.

Generally, motion distortions are more serious in non-repetitive scanning mode, which brings the extra difficulty for point registration. Lin and Zhang~\cite{lin2020loam} segment a single scan into several pieces to reduce the motion distortions. Xu and Zhang~\cite{xu2021fast} de-skew the points by propagating the relative pose back from IMU measurements. In our pipeline, we employ the continuous-time model like~\cite{dellenbach2022ct} in optimization, which is capable of undistorting the raw points in a scan without extra sensors.

\begin{figure*}[t]
	\centering
	\includegraphics[width=\textwidth]{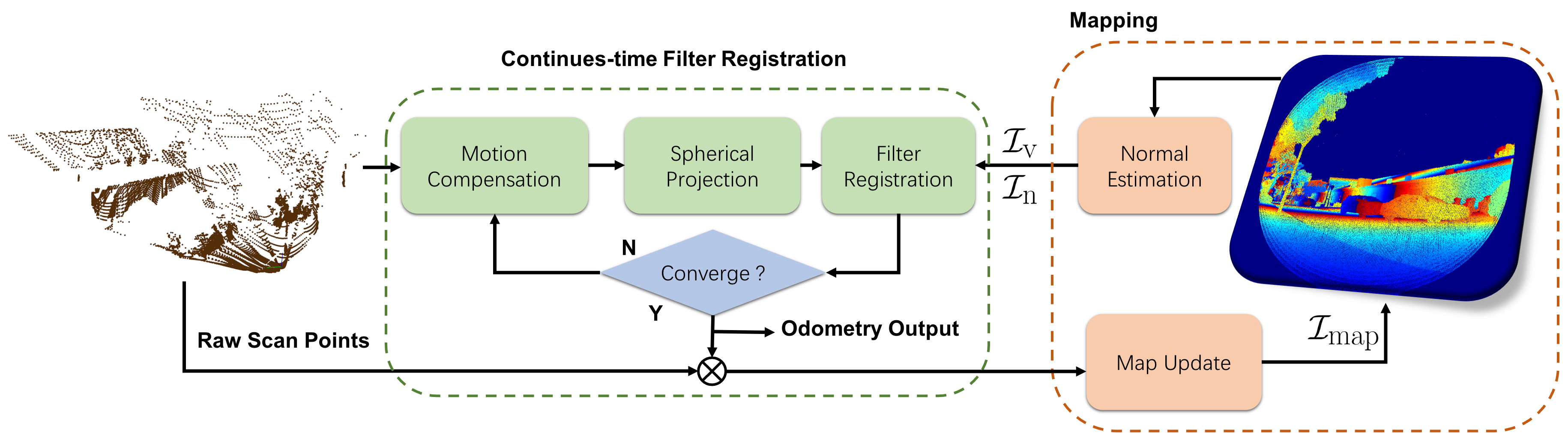}
	\caption{\footnotesize Overview of our proposed ECTLO approach. Once the scan is obtained, the optimal state $\mathbf{s}=[\mathbf{T}_{\mathrm{WL}_{b}},\mathbf{T}_{\mathrm{WL}_{e}}]$ is predicted by continuous-time filter registration. The mapping module combines the input scan $\mathcal{P}$ with the previous range map by the odometry result $\mathbf{s}$ to form an updated range image $\mathcal{I}_{\textrm{map}}$. The memory consumption of $\mathcal{I}_{\textrm{map}}$ is fixed depending on the user-defined FoV and angular resolution of image.}
	\label{fig:pipeline}
	\vspace{-0.2in}
\end{figure*}

\subsection{Point Set Registration}
Generally, LiDAR odometry is treated as a point-set registration problem in literature. Given the input point clouds perceived by LiDAR at two consecutive timestamps, Iterative Closed Point (ICP) algorithm~\cite{besl1992method} is the typical solution to align them, which updates the relative transformation iteratively until convergence. LOAM~\cite{zhang2014loam} is the most successful LiDAR odometry approach, which selects the feature points by computing the roughness of scattered points on each scan line. Moreover, a variant of ICP algorithm is employed to make use of both point-to-line~\cite{censi2008icp} and point-to-plane loss~\cite{low2004linear}, which achieves low-drift odometry with real-time performance.

The accuracy of ICP registration depends on the reliable correspondences between point sets, which is suspect to noise, outliers, and occlusions~\cite{myronenko2010point}. To this end, the probabilistic model is usually regarded as a more robust approach to align point sets. Generalized-ICP~\cite{segal2009generalized} formulates the point set registration problem into a probabilistic framework, which unifies point-to-point, point-to-plane and plane-to-plane ICP by covariance estimation. Similarly, Normal Distribution Transformation (NDT)~\cite{magnusson2007scan} replaces the single point by normal distribution within a pre-defined regular voxel. These correlation-based registration methods can be interpreted as minimizing the distance between distributions. 

Assuming that each point has a Gaussian variance, point cloud can be regarded as a Gaussian Mixture Model (GMM)~\cite{jian2010robust,myronenko2010point}. However, GMM-based methods are computationally intensive, which hinder them from real-time applications. Gao and Tedrake~\cite{gao2019filterreg} present a probabilistic model, which computes the filter-based correspondences in E step and updates pose by Gauss-Newton algorithm in M step. In this paper, we employ filter registration to tackle the problem from image map representation, in which  the permutohedral filter is replaced by an efficient patch filter on range image~\cite{zheng2021efficient}.

\section{METHODOLOGY}
In this section, we present the details of our proposed odometry approach. Firstly, we use a single 2D range image for map representation. Then, we suggest a continuous-time FilterReg method to achieve robust registration and compensate the severe motion distortions in cost function. Moreover, the analytic Jacobians are derived for efficient optimization. Fig.~\ref{fig:pipeline} shows the overview of our pipeline.

\subsection{Preliminary}
Assume that the point measurement is relative to LiDAR coordinate system $\textrm{L}_{t}$ at timestamp $t$. We denote the orientation and position of LiDAR as  $\mathbf{T}_{\textrm{WL}_{t}}$ with respect to the world frame $\textrm{W}$ at timestamp $t$, where the world frame is $\textrm{W}=\textrm{L}_{0}$ at the starting location.

In this paper, the movement of LiDAR is in special Euclidean space. To obtain an unconstrained minimization problem, we apply the twist parameterization from Lie Algebra~\cite{sola2018micro}, where a vector $\boldsymbol{\xi} \in \mathbb{R}^{6}$ represents rigid transformation $\mathbf{T}_{\mathrm{WL}} \in \mathrm{SE(3)}$. All operators are right-version as below
\begin{equation}
\begin{aligned}
     \textup{right-}\oplus  :\mathcal{Y} = \mathcal{X}\oplus \boldsymbol{\xi} = \mathcal{X} \mathrm{Exp}(\boldsymbol{\xi}) \in \mathrm{SE(3)} \\
    \textup{right-}\ominus :\boldsymbol{\xi} =\mathcal{Y} \ominus \mathcal{X} = \mathrm{Log}(\mathcal{X}^{-1} \mathcal{Y})\in\mathbb{R}^{6} .
\end{aligned}
\end{equation}
where $\mathcal{X},\mathcal{Y}\in \mathrm{SE(3)}$. $\mathrm{Log}(\cdot):\mathrm{SE(3)} \rightarrow \mathbb{R}^{6}$ is the logarithmic mapping, and $\mathrm{Exp}(\cdot):\mathbb{R}^{6} \rightarrow \mathrm{SE(3)}$ is its inverse exponential mapping.

\subsection{Range Image Map Representation}\label{sec:estep}
As a registration problem, the pose estimation results are dependent on the overlap region of consecutive scans. Since the point cloud from multi-line spinning LiDAR has a 360-degree horizontal FoV, previous methods~\cite{lin2020loam,xu2021fast,bai2022faster} keep a local map that stores the historical points around current center with a fixed radius to achieve better scan-to-map accuracy. However, the prism-based LiDAR has a limited horizontal FoV as the camera. The region of interest is smaller than spinning LiDAR. While we have the option to choose points within LiDAR's FoV and maintain KD-Tree~\cite{zhang2014loam,shan2018lego} or voxel~\cite{magnusson2007scan} map representations, the range image~\cite{zheng2021efficient} that retains points within FoV proves to be a more computationally efficient data structure with low memory consumption.

We extend this representation from the conventional multi-line spinning LiDARs~\cite{zheng2021efficient} to small FoV LiDAR.

\subsubsection{Spherical Projection}
A range image is an index table $\mathcal{I}:\mathbb{R}^{2} \rightarrow \mathbb{R}^{3}$ that reserves the spatial relationship of 3D point set within single 2D image. Given a point $(x,y,z)^{\top}$ relative to range image origin, its correspondence can be indexed in previous image by spherical projection $\Pi:\mathbb{R}^{3} \rightarrow \mathbb{R}^{2}$
\begin{equation}
\label{equ:project}
    \begin{bmatrix}
u
\\ 
v
\end{bmatrix}=\Pi(x,y,z)=
\begin{bmatrix}
\left ( 1/2+\arctan \left ( y/x \right ) \cdot f^{-1}_{h} \right )w
\\ 
\left ( 1/2-\arcsin\left ( z/r \right )  \cdot f^{-1}_{v} \right )h
\end{bmatrix},
\end{equation}
where $r=\sqrt{x^{2}+y^{2}+z^{2}}$. The image center correspondence to the current LiDAR center. $w$ and $h$ are the width and height of image. The pixel length of $w$ and $h$ is proportional to angular resolution $\beta$ and FoV, where $w=\beta f_{h}$ and $h=\beta f_{v}$. Note that $f_{h}$ and $f_{v}$ are the user-defined FoV rather than the original LiDAR's, which keeps the historical points out of the sensor's FoV to facilitate robust registration.   


\subsubsection{Map Generation and Updating}
The generation of initial range image $\mathcal{I}_{\textrm{map}}$ is just the process that project points onto an empty image. Fig.~\ref{fig:pattern} shows the single scan projection results of different LiDARs. For Livox Mid-40, its single scan is too sparse to register, several consecutive scans at the static position are used for initialization. If the projection pixel is already occupied by another point, the collision is resolved by selecting the nearest points. In our proposed approach, we only maintain single range image $\mathcal{I}_{\textrm{map}}$. This means that the memory consumption of map is fixed when the FoV and angular resolution of the range image are determined. 

Range Image $\mathcal{I}_{\textrm{map}}$ is a robocentric map representation, where the projection origin is the last successful registered LiDAR position. Once current scan can register with previous $\mathcal{I}_{\textrm{map}}$, we transfer map points to the local scan coordinate. Then, all the previous map points and current scan points are projected onto a new image. We choose the nearest points when pixels conflict. This new image $\mathcal{I}_{\textrm{map}}$ is the local map for future registration.

However, occlusion is inevitable when keeping historical map points in range image representation even with complicated conflict-resolving solutions. This phenomenon makes our $\mathcal{I}_{\textrm{map}}$ noisier after continuously updating. Since the odometry output of current scan is affected by the local map, it may lead to the inferior pose optimization results. We handle this issue by applying an efficient GMM-based registration method.

\subsection{Continuous-time Filter Registration on Range Image}

In general, robocentric range image usually reflects the surface environment in LiDAR's FoV so that the noisy data may be injected into this 2D map due to occlusions after updating. Therefore, the conventional registration methods like ICP and its variants~\cite{zhang2014loam,lin2020loam,behley2018efficient} may not be effective for this map representation. Another common issue of LiDAR is motion distortion. Motivated by the probabilistic method FilterReg~\cite{gao2019filterreg}, we extend it into a continuous-time formulation, which is more robust to noise, outliers and occlusions. Moreover, a Gaussian filtering algorithm is employed to efficiently find the correspondences on 2D range image, where the analytic Jacobians are derived for Gauss-Newton optimization.

\subsubsection{Odometry Formulation}
The fundamental problem of LiDAR odometry is to find a series of discrete motion parameters properly describing the LiDAR movement. Once the raw scan $\mathcal{P}$ is obtained, the transformation between the current scan $\mathcal{P}$ and previous map $\mathcal{Q}$ is estimated. $\mathbf{p}_{1},\mathbf{p}_{2},\cdots,\mathbf{p}_{M}$ and $\mathbf{q}_{1},\mathbf{q}_{2},\cdots,\mathbf{q}_{N}$ are points in $\mathcal{P}$ and $\mathcal{Q}$, respectively.

We make use of a scan-to-map method, where map $\mathcal{Q}$ stores in $\mathcal{I}_{\textrm{map}}$. With the continuous updating of the range map, it becomes quite noisy. Conventional ICP methods aim to find the exact closest point in the data association procedure, which is susceptible to noise data. To this end, we introduce an efficient GMM-based method  FilterReg~\cite{gao2019filterreg} into our pipeline. To achieve better accuracy in optimization, we adopt point-to-plane criteria. The original FilterReg estimates the current state $\mathbf{s}$ of LiDAR using the EM algorithm, as follows:

\textbf{E step}: for each point in scan $\mathcal{P}$, compute
\begin{equation}
\begin{aligned}
    \mathbf{m}^{0}_{\mathbf{p}_{i}} &=\sum_{\mathbf{q}_{j}}\mathcal{N}(\mathbf{p}_{i}(\mathbf{s}^{\text{old}});\mathbf{q}_{j},\Sigma_{xyz} ) \\
    \mathbf{m}^{1}_{\mathbf{p}_{i}} &=\sum_{\mathbf{q}_{j}}\mathcal{N}(\mathbf{p}_{i}(\mathbf{s}^{\text{old}});\mathbf{q}_{j},\Sigma_{xyz} )\mathbf{q}_{j} \\
    \mathbf{n}_{\mathbf{p}_{i}} &=(\sum_{\mathbf{q}_{j}}\mathcal{N}(\mathbf{p}_{i}(\mathbf{s}^{\text{old}});\mathbf{q}_{j},\Sigma_{xyz} )\mathbf{n}_{\mathbf{q}_{j}})/\mathbf{m}^{0}_{\mathbf{p}_{i}}
\end{aligned}
\end{equation}

\textbf{M step}: minimize the following objective function
\begin{equation}
\label{equ:mstep}
    E_{\textrm{reg}}(\mathbf{s}) = \frac{1}{M} \sum_{\mathbf{p}_{i}}\frac{\mathbf{m}^{0}_{\mathbf{p}_{i}}}{\mathbf{m}^{0}_{\mathbf{p}_{i}}+c}\text{dot}\left (\mathbf{n}_{\mathbf{p}_{i}},\mathbf{p}_{i}(\mathbf{s})-\frac{\mathbf{m}^{1}_{\mathbf{p}_{i}}}{\mathbf{m}^{0}_{\mathbf{p}_{i}}}\right)^{2},
\end{equation}
where $c=\frac{w_{i}}{1-w_{i}}\frac{J}{M}$, and $w_{i}$ is the parameter that accounts for the ratio of outliers. $\mathcal{N}(\cdot)$ is Gaussian distribution. Gaussian kernel $\Sigma_{xyz}=\textup{diag}(\sigma^{2},\sigma^{2},\sigma^{2})$ is the fixed parameter, and the variance $\sigma$ decides the weight of each map points $\mathbf{q}_{j}$ and its normal vector $\mathbf{n}_{\mathbf{q}_{j}}$. $\mathbf{p}_{i}(\mathbf{s})$ transfers scan points from local coordinate into world frame, which may extend to continuous-time form. Practically, it is inefficient to employ all map points in E step. Thus, we select the neighborhood points of $\mathbf{p}_{i}(\mathbf{s})$ in a window for approximation, where $J$ is the size of valid neighborhoods. The implementation details are following.

\subsubsection{Implementation on Range Image}
Just like the closet point searching in ICP, the E step is to compute the correspondence of scan point $\mathbf{p}_{i}$. The form of $\mathbf{m}^{0}_{\mathbf{p}_{i}},\mathbf{m}^{1}_{\mathbf{p}_{i}},\mathbf{n}_{\mathbf{p}_{i}}$ are Gaussian Transform. Obviously, the bottleneck of this EM algorithm is how to efficiently compute thousands of the above independent items in E step.  FilterReg~\cite{gao2019filterreg} employs a customized permutohedral lattice filter to enable efficient computation without compromising accuracy. Since the point registration problem occurs in 3D space rather than the general N-dimensional space, we utilize a more efficient 2D Gaussian filter on the range image.

The advantage of range image is to retain the 3D spatial relationship within a 2D index table, where its neighbor exists in adjacent pixels. Therefore, the Gaussian Transform can be efficiently computed by image filter methods like Gaussian Blur or Bilateral Filter. If the spherical projection result of $\mathbf{p}_{i}$ is $(u_{i},v_{i})$, the related map points affecting the correspondence are in adjacent pixels. Hereby, Gaussian filter is employed to compute the filter-based correspondence $\mathbf{m}^{0}_{\mathbf{p}_{i}},\mathbf{m}^{1}_{\mathbf{p}_{i}},\mathbf{n}_{\mathbf{p}_{i}}$ in a pre-defined window $\mathcal{W}_{i}$ with center $(u_{i},v_{i})$. 

 During scan registration, map points $\mathcal{Q}$ are fixed so that their normal vectors do not change. By making use of point-to-plane criteria, we convert the previous map $\mathcal{I}_{\textrm{map}}$ into two temporal index tables, vertex map $\mathcal{I}_{\textrm{v}}$ storing 3D position information, and normal map $\mathcal{I}_{\textrm{n}}$ saving their normal vector. We estimate normals through Eigendecomposition within a $\mathcal{W}{n}$ window on the range image. To select valid planar points, we consider those surface curvatures~\cite{zheng2021efficient} that fall below the threshold $\delta{\sigma}$. In cases where the angle between the normal vector and the point exceeds $90^{\circ}$, we flip the normal direction. It is important to note that we only need to calculate the normals once before the EM procedure.

\subsubsection{Motion Compensation}\label{sec:ct}
A scan is the accumulated point cloud during a period of time $t\in \left [ t_{b},t_{e} \right )$. $t_{b}$ is the starting timestamp of a scan, and $t_{e}$ is the end timestamp. Motion distortion occurs due to assuming that all points are sampled simultaneously. To address this issue, we employ a continuous-time model to compensate the motion distortions in filter registration.

The continuous movement of LiDAR within a scan is parameterized by the beginning of scan $\mathbf{T}_{\textrm{WL}_{b}}$ and end of scan $\mathbf{T}_{\textrm{WL}_{e}}$. Therefore, the estimation target is the state
$\mathbf{s}=[\mathbf{T}_{\textrm{WL}_{b}},\mathbf{T}_{\textrm{WL}_{e}}]$, where $\mathbf{s}$ is in $\mathbb{R}^{12}$ through the minimal representation of Lie algebra. Given a raw point measurement $\mathbf{p}_{i} \in \mathbb{R}^{3}$ at timestamp $t_{i} \in \left [ t_{b},t_{e} \right )$, its corrected position $\mathbf{p}_{i}(\mathbf{s})$ in world coordinate is computed by
\begin{equation}
\begin{aligned}
    \boldsymbol{\tau} &=\mathbf{T}_{\mathrm{WL}_{e}}\ominus \mathbf{T}_{\mathrm{WL}_{b}}  \\
    \alpha &=(t_{i}-t_{b})/(t_{e}-t_{b}) \\
    \mathbf{T}_{\mathrm{WL}_{i}} &=\mathbf{T}_{\mathrm{WL}_{b}}\mathrm{Exp}(\alpha \boldsymbol{\tau} ) \\
    \mathbf{p}_{i}(\mathbf{s}) &=\mathbf{T}_{\mathrm{WL}_{i}}\mathbf{p}_{i}
\end{aligned}
\end{equation}
where $\boldsymbol{\tau} \in \mathbb{R}^{6}$ is the tangent space of manifold $\mathrm{SE(3)}$, and $\mathbf{T}_{\mathrm{WL}_{i}}$ is the LiDAR origin at timestamp $t_{i}$ in world coordinate. Each 3D point $(x,y,z)^{\top}$ is represented in homogeneous coordinates as $\mathbf{p}_{i}=(x,y,z,1)^{\top}$. As the estimated state $\mathbf{s}$ is updated iteratively until the convergence, the point $\mathbf{p}_{i}(\mathbf{s})$ is recomputed with the newly updated pose at each iteration.  

Ideally, the beginning pose of current scan should be consistent with the end pose of previous one. Thus, the new state is initialized by its previous pose as follows
\begin{equation}
\begin{aligned}
    \mathbf{T}_{\mathrm{WL}_{b}^{n}}&=\mathbf{T}_{\mathrm{WL}_{e}^{n-1}} \\
    \mathbf{T}_{\mathrm{WL}_{e}^{n}}&=\mathbf{T}_{\mathrm{WL}_{b}^{n}}(\mathbf{T}_{\mathrm{WL}_{b}^{n-1}})^{-1}\mathbf{T}_{\mathrm{WL}_{e}^{n-1}},
\end{aligned}
\end{equation}
where the first state $\mathbf{s}_{0}$ is set to identity. However, the optimization may diverge on the noisy data with strict consistency constraints. Therefore, we impose two soft constraints, the location consistency $E_{\textrm{loc}}(\mathbf{s}) = \mathbf{r}_{\textup{loc}}^{\top}\mathbf{r}_{\textup{loc}}$ and velocity consistency $E_{\textrm{vel}}(\mathbf{s})=\mathbf{r}_{\textup{vel}}^{\top}\mathbf{r}_{\textup{vel}}$, where the position difference is defined as $\mathbf{r}_{\textup{loc}}  =\mathbf{T}_{\mathrm{WL}_{b}^{n}}\ominus \mathbf{T}_{\mathrm{WL}_{e}^{n-1}} 
$ and velocity difference between the consecutive scans is computed by $    \mathbf{r}_{\textup{vel}} =(\mathbf{T}_{\mathrm{WL}_{e}^{n}}\ominus \mathbf{T}_{\mathrm{WL}_{b}^{n}}) - (\mathbf{T}_{\mathrm{WL}_{e}^{n-1}}\ominus \mathbf{T}_{\mathrm{WL}_{b}^{n-1}})$. Since handheld devices exhibit complex and irregular movement compared to driving scenarios, their scan data contain significant noise. Thus we apply the state constraint in the $\mathrm{SE}(3)$ space, rather than solely the translation component~\cite{dellenbach2022ct}.

With the above proposed motion constraints, we reformulate the previous M step in Eqn.~\ref{equ:mstep} into the continuous-time form,
\begin{equation}
   E_{\textrm{ct-reg}}(\mathbf{s})=E_{\textrm{reg}}(\mathbf{s})+\lambda_{l}E_{\textrm{loc}}(\mathbf{s})+\lambda_{v}E_{\textrm{vel}}(\mathbf{s}),
\end{equation}
where $\lambda_{l}$ and  $\lambda_{v}$ are two regularization coefficients to balance the above the energy terms.

\subsubsection{Gauss-Newton in EM Procedure}\label{sec:jacobian}

The optimization in the M step involves a least square minimization problem, which can be reformulated into the generalized form $\sum c_{i}\mathbf{r}{i}^{\top}\mathbf{r}{i}$. Here, $c_{i}$ represents the weight coefficient, and $\mathbf{r}_{i}$ denotes the residual of the energy term. In our proposed approach, there are three kinds of residual, including $\mathbf{r}_{\textup{reg}}^{i}=\mathbf{n}_{\mathbf{p}_{i}}^{\top }(\mathbf{p}_{i}(\mathbf{s})-\mathbf{m}^{1}_{\mathbf{p}_{i}}/\mathbf{m}^{0}_{\mathbf{p}_{i}})$, $\mathbf{r}_{\textup{loc}}$ and $\mathbf{r}_{\textup{vel}}$. By the first order Taylor expansion, the residual around $\mathbf{s}$ is approximated as follows
\begin{equation}
    \mathbf{r}_{i}(\mathbf{s}+\Delta \mathbf{s})\simeq \mathbf{r}_{i}(\mathbf{s})+\frac{\partial \mathbf{r}_{i}}{\partial \mathbf{s}}\Delta \mathbf{s},
\end{equation}
where $\frac{\partial \mathbf{r}_{i}}{\partial \mathbf{s}}$ is the Jacobian, and $\Delta \mathbf{s}=\begin{bmatrix}\Delta \boldsymbol{\xi}_{b} & \Delta\boldsymbol{\xi}_{e} \end{bmatrix}$ is the increment. In Gauss-Newton (GN) algorithm, the optimal increment minimizing $E_{\textrm{ct-reg}}(\mathbf{s})$ is found by solving the linear equation $\mathbf{H}\Delta \mathbf{s}=-\mathbf{b}$, where the Hessian matrix $\mathbf{H}=\sum c_{i}\left ( \frac{\partial \mathbf{r}_{i}}{\partial \mathbf{s}} \right )^{\top }\frac{\partial \mathbf{r}_{i}}{\partial \mathbf{s}}$ and $\mathbf{b}=\sum c_{i}\left ( \frac{\partial \mathbf{r}_{i}}{\partial \mathbf{s}} \right )^{\top }\mathbf{r}_{i}$.

Since $\mathbf{T}_{\mathrm{WL}_{b}}$ and $\mathbf{T}_{\mathrm{WL}_{e}}$ are represented in $\mathrm{SE(3)}$, the new state $\mathbf{s}$ for the next E step is updated through $\mathbf{s}=\mathbf{s}^{\textrm{old}} \oplus \Delta \mathbf{s}$. Specifically, the two poses are updated as follows
\begin{equation}
    \mathbf{T}_{\mathrm{WL}_{b}}\leftarrow  \mathbf{T}_{\mathrm{WL}_{b}} \oplus \Delta \boldsymbol{\xi}_{b}, \mathbf{T}_{\mathrm{WL}_{e}}\leftarrow \mathbf{T}_{\mathrm{WL}_{e}} \oplus \Delta \boldsymbol{\xi}_{e}.
\end{equation}
The EM procedure will terminate when either the maximum increment falls below the threshold $\delta_{m}$ or the number of iterations exceeds 15.

The computational cost in M step is mainly dominated in calculating the Jacobian. To this end, we derive the analytic Jacobians of each residual with respect to the beginning pose $\mathbf{T}_{\textrm{WL}_{b}}$ and the end pose $\mathbf{T}_{\textrm{WL}_{e}}$ for efficient optimization. The analytic Jacobian of registration term is computed by
\begin{equation}
\begin{aligned}
    \frac{\partial \mathbf{r}_{\textup{reg}}^{i}}{\partial \mathbf{s}} &= \mathbf{n}_{\mathbf{p}_{i}}^{\top }\frac{\partial (\mathbf{T}_{\mathrm{WL}_{i}}\mathbf{p}_{i})}{\partial \mathbf{T}_{\mathrm{WL}_{i}}}
    \begin{bmatrix}
    \frac{\partial \mathbf{T}_{\mathrm{WL}_{i}}}{\partial \mathbf{T}_{\mathrm{WL}_{b}}} & 
    \frac{\partial \mathbf{T}_{\mathrm{WL}_{i}}}{\partial \mathbf{T}_{\mathrm{WL}_{e}}}
    \end{bmatrix}  \\
    \frac{\partial (\mathbf{T}_{\mathrm{WL}_{i}}\mathbf{p}_{i})}{\partial \mathbf{T}_{\mathrm{WL}_{i}}} &= 
    \begin{bmatrix}
    \mathbf{R}_{\mathrm{WL}_{i}}  & -\mathbf{R}_{\mathrm{WL}_{i}}\left [\mathbf{p}_{i}  \right ]_{\times }
    \end{bmatrix}  \\
    \frac{\partial \mathbf{T}_{\mathrm{WL}_{i}}}{\partial \mathbf{T}_{\mathrm{WL}_{b}}} &= (1-\alpha)\mathbf{J}_{r}((\alpha-1)\boldsymbol{\tau})\mathbf{J}_{l}^{-1}(\boldsymbol{\tau})\\
\frac{\partial \mathbf{T}_{\mathrm{WL}_{i}}}{\partial \mathbf{T}_{\mathrm{WL}_{e}}} &= \alpha\mathbf{J}_{r}(\alpha\boldsymbol{\tau})\mathbf{J}_{r}^{-1}(\boldsymbol{\tau}),
\end{aligned}
\end{equation}
As in~\cite{sola2018micro}, $\mathbf{J}_{r}(\cdot)$ and $\mathbf{J}_{l}(\cdot)$ are the right- and left- Jacobians, respectively. $\left [\cdot  \right ]_{\times }$ denotes the skew symmetric matrix. 

Similarly, the analytic Jacobians of motion constraints can be derived as below 
\begin{equation}
    \begin{aligned}
        \frac{\partial \mathbf{r}_{\textup{loc}}}{\partial \mathbf{s}} &=\begin{bmatrix}
\mathbf{J}_{r}^{-1}(\mathbf{r}_{\textup{loc}}) & \mathbf{0}
\end{bmatrix} \\
\frac{\partial \mathbf{r}_{\textup{vel}}}{\partial \mathbf{s}} &=\begin{bmatrix}
-\mathbf{J}_{l}^{-1}(\boldsymbol{\tau}) & \mathbf{J}_{r}^{-1}(\boldsymbol{\tau})
\end{bmatrix}.
    \end{aligned}
\end{equation}

\section{EXPERIMENT}
In this section, we present the details of our experiments and discuss the results of LiDAR odometry. We evaluate our proposed approach on several datasets collected by various prism-based LiDAR with different scanning patterns. An ablation study proves the effectiveness of our continuous-time filter registration module. Additionally, we present the efficiency analysis on both commodity laptop and the embedded devices. Our method is implemented in C++ with CUDA. The experiments are performed on a laptop computer with an Intel Core i7-9750H CPU@2.60 GHz having 16GB RAM and an NVIDIA GeForce RTX 2060 GPU. Besides, we evaluate the computational time of our proposed approach on NVIDIA Jetson AGX, which are the popular embedded devices in robotics. For simplicity, we name our method as `ECTLO'. Table~\ref{tab:parameters} gives the parameter settings. 

\begin{table}[t]
    \centering
        \caption{Parameter Settings of Our Method}
    \begin{threeparttable}
    \begin{tabular}{@{}cc|cc@{}}
    \toprule
     Parameter & value &  Parameter & value\\
     \midrule
        $f_{h},f_{v}$ (Livox Mid-40) & $50^{\circ},50^{\circ}$ & $\lambda_{l},\lambda_{v}$ & 0.1, 0.1  \\
        $f_{h},f_{v}$ (Livox Horizon) & $90^{\circ},30^{\circ}$ &  $\sigma$ & 0.25 \\
        $f_{h},f_{v}$ (Livox Avia) & $80^{\circ},80^{\circ}$ & $w_{i}$ & 0.2\\
        $f_{h},f_{v}$ (Livox Mid-70) & $80^{\circ},80^{\circ}$ & $\beta$ & 10 $\textup{pixels}/\circ$ \\
        $\mathcal{W}_{i}$  & $7\times 7$ & $\delta_{\sigma}$ & 0.055\\
        $\mathcal{W}_{n}$  & $5\times 5$ & $\delta_{m}$ & 5e-4\\
    \bottomrule
    \end{tabular}
    \end{threeparttable}
    \label{tab:parameters}
    \vspace{-0.2in}
\end{table}

\subsection{Performance Evaluation }

\begin{figure*}[t]
    \centering
    \includegraphics[width=\textwidth]{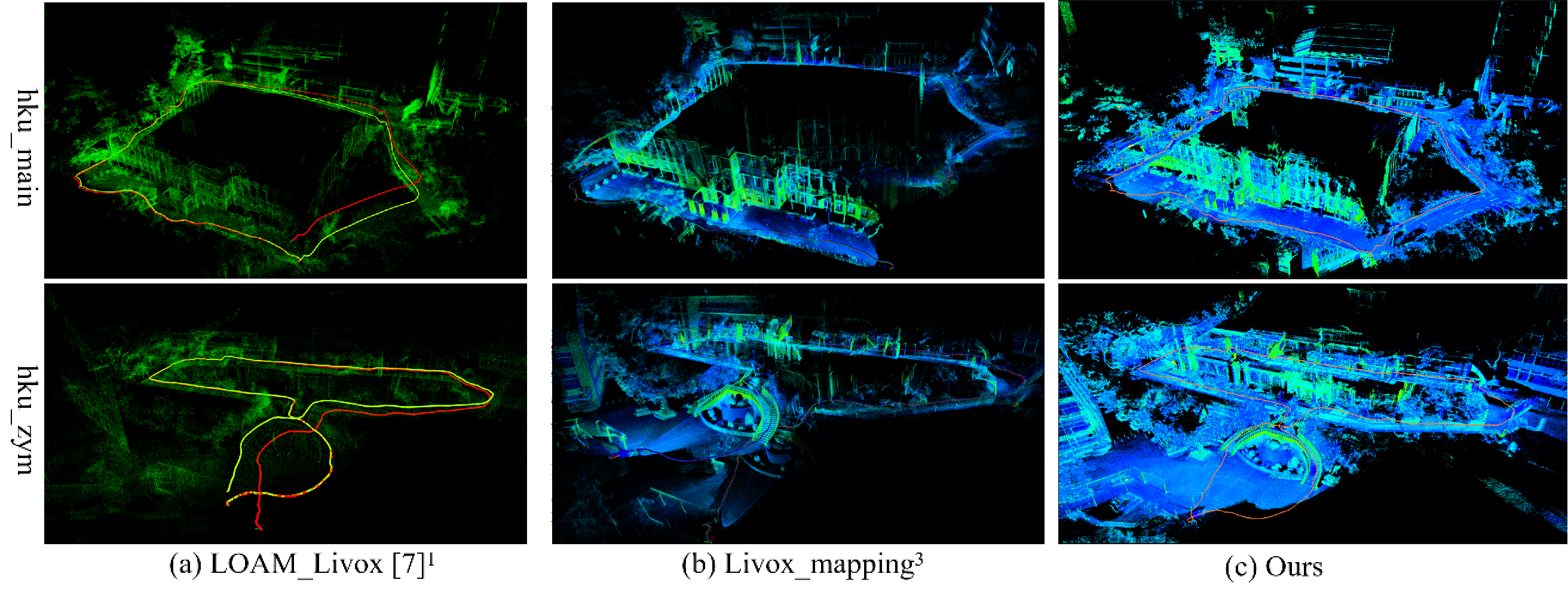}
    \vspace{-0.2in}
    \caption{\footnotesize The trajectory and reconstruction results of different methods on public Livox Mid-40 dataset. The green line in (a) is the trajectory with an extra loop closure module. The red line indicates the result without loop closure.}
    \label{fig:eva_mid40}
\end{figure*}
We compare our presented ECTLO method against the state-of-the-art LiDAR approaches with publicly available implementations, including LOAM-Livox~\cite{lin2020loam}\footnote{https://github.com/hku-mars/loam\_livox}, BALM~\cite{liu2021balm}\footnote{https://github.com/hku-mars/BALM}, and Livox-mapping\footnote{https://github.com/Livox-SDK/livox\_mapping}. 
\subsubsection{Qualitative Comparison}
The testing data is adopted from the previous studies~\cite{lin2020loam} with a Livox Mid-40. Since the dataset does not have ground truth for validation, we select two sequences `loop\_hku\_main' (389m) and `loop\_hku\_zym' (290m) whose beginning and end positions are the same. 

Fig.~\ref{fig:eva_mid40} plots the trajectories of each valid method. It can be clearly seen that our proposed  ECTLO approach achieves the lowest drift even without the extra loop closure module, which obtains the high reconstruction quality by taking advantage of the accurate continuous-time filter registration. Livox-mapping is the official odometry package from Livox, however, its mapping results have the obvious drifts. It is worthy of mentioning that BALM totally fails to obtain the correct results with this small FoV LiDAR.

\begin{figure}[t]
	\centering
	\includegraphics[width=0.5\textwidth]{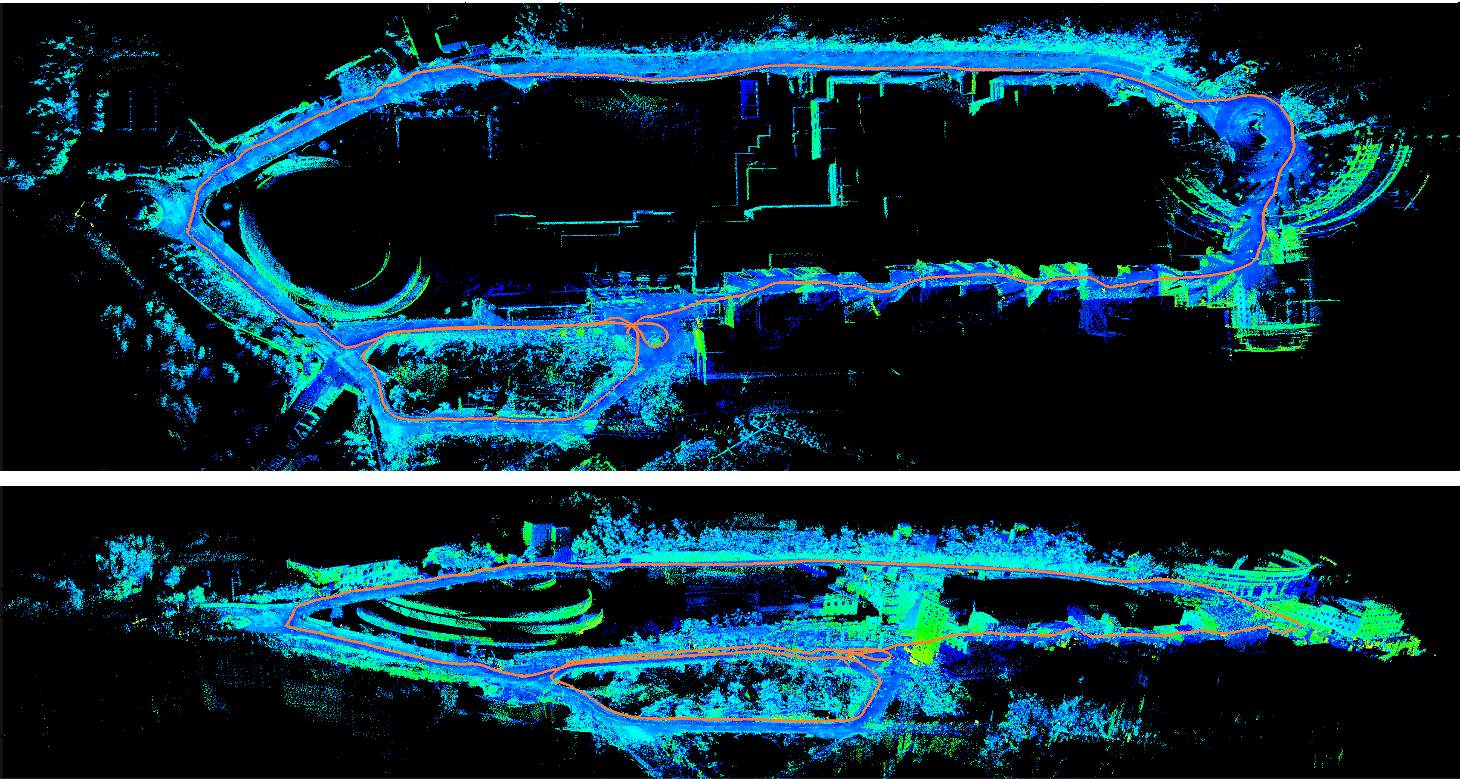}
	\caption{Our mapping result on large-scale outdoor sequence `hkust\_campus\_01'.}
	\label{fig:hkust01}
	\vspace{-0.25in}
\end{figure}

To examine the gap between LiDAR-only odometry and sensor fusion methods, we compare our proposed LiDAR only approach against the state-of-the-art LiDAR-inertial odometry FAST-LIO2~\cite{xu2021fast} in a large-scale outdoor sequence `hkust\_campus\_01' (1524m) from R3LIVE~\cite{lin2022r} collected by Livox Avia. The end-to-end drift of ECTLO is 4.72m in contrast to 0.31m drift of FAST-LIO2\footnote{https://github.com/hku-mars/FAST\_LIO}. LOAM\_Livox cannot generalize to Livox Avia, and Livox\_mapping has large drift. Moreover, we plot the mapping result in Fig.~\ref{fig:hkust01}.

\subsubsection{Quantitative Comparison}
The Hilti SLAM Challenge Dataset~\cite{helmberger2022hilti} is a multi-sensor collection containing challenging featureless areas and varying illumination conditions. We evaluate different odometry by computing absolute trajectory error (ATE)~\cite{grupp2017evo} using Livox Mid-70 data. Meanwhile, SVO2~\cite{forster2016svo} and hdl\_graph\_slam~\cite{koide2019portable} are two baselines of VIO and multi-line spinning LiDAR odometry, respectively. The Mid-70 Data in Lab sequence is invalid at the beginning where the sensor is too close to the wall. Thus, we select other five sequences with ground truth for evaluation. 

\begin{table*}[t]
\centering
\caption{ATE (m) on the Hilti SLAM Challenge Dataset}
\begin{threeparttable}
\begin{tabular}{@{}cc|ccccc|c@{}}
\toprule
 & Sensor& RPG Area 
  & Basement1
 & Basement4
 & Construction2 & Campus2 & 
Average ATE \\
\midrule
 SVO~\cite{forster2016svo} &Camera + IMU& 1.927 &  0.815 &2.609 & 2.986 & 8.948 & 3.618\\
 \midrule
 hdl\_graph\_slam~\cite{koide2019portable} & Ouster-64
 & 0.350 &  \bf{0.279} &0.336 & 0.741 & 0.353 & 0.445\\
\midrule
Livox\_mapping &Mid-70  &  6.043 & ×\tnote{1}  &1.988 & 15.479 & 10.924 & -\\

\midrule
Fast-LIO2~\cite{xu2021fast}&Mid-70 + IMU &\bf{0.242} &2.334 & 1.407 & \bf{0.206} & \bf{0.091} & 0.856\\

   \midrule
ECTLO &  Mid-70 &0.394 & 0.334&\bf{0.124} & 0.253 & 0.356 & \bf{0.282}\\
\bottomrule
\end{tabular}

\begin{tablenotes}
\item[1] Fail registration in this sequence.
\end{tablenotes}

\end{threeparttable}
\label{tab:ate}
\end{table*}

Table~\ref{tab:ate} shows the ATE on Hilti SLAM Challenge dataset, where LOAM\_Livox and BALM fail to track on all sequences. Sequence `RPG Area' and `Basement1' have the degenerated scenarios, where data is recorded in front of a textureless wall for a long time. It is challenging for both camera and LiDAR with small FoV. It can be observed that ECTLO outperforms SVO using the visual-inertial sensor and Livox\_mapping using Livox Mid-70 at a large margin. Our result is even close to the method using 360$^{\circ}$ scanning LiDAR Ouster OS0-64. 

Although FAST-LIO2 using Livox Mid-70 and IMU has better performance at outdoor sequences, it drifts significantly in the indoor basement even with an extra sensor. This reflects that our robocentric image representation is a more appropriate structure for LiDAR with small FoV in narrow space. Overall, our ECTLO achieves the lowest average ATE on the Hilti Dataset with single small FoV LiDAR.

\subsection{Ablation Study}
The main contributions of our proposed method are continuous-time model for motion compensation and GMM-based filter registration for LiDAR odometry. To investigate the effectiveness of each module, we conduct an ablation study with various settings. The experiments are performed on three sequences with different scanning patterns, including private `ZJU-M0', `ZJU-H0' and public `hku\_park\_00'. The private datasets are collected by ourselves, which are publicly available\footnote{https://github.com/kevin2431/ECTLO}. Table~\ref{tab:ablation} shows the end-to-end errors of different combinations. It can be observed that the continuous-time filter registration significantly improves the odometry accuracy. Although continuous time module performs well on the easy sequences like `ZJU-M0' and `hku\_park\_00', it fails on the challenging `ZJU-H0' sequence. It can be concluded that both continuous-time and GMM-based registration are essential to achieving the robust and accurate odometry results in various scenarios.

To fairly evaluate each module, we compute the ATE on three sequences of Hilti Dataset. Sequence `RPG Area' is in middle size indoor space. Sequence `Basement4' has low-speed motion in the structural environment and `Construction2' has fast motion in the complex outdoor large-scale scenario. Continuous-time filter registration achieves the lowest ATE in all sequences.

\begin{table*}[t]
\centering
\caption{Ablation Study on Our Proposed Approach.}
\label{tab:ablation}

\begin{threeparttable}
\begin{tabular}{@{}ccccc|ccc@{}}
\toprule
CT Module & GMM Registration & \multicolumn{3}{@{}c@{}}{End-to-end Error (m)} &
\multicolumn{3}{@{}c@{}}{ATE (m)}\\
\midrule

& & ZJU-M0 &ZJU-H0 & hku\_park\_00 
&RPG Area &Basement4 & Construction2 \\
\cmidrule{3-8}

 \XSolidBrush\tnote{1} & \XSolidBrush\tnote{2} & ×\tnote{3} & × & 37.02& 17.32 & 7.67& ×\\ 
  \XSolidBrush & \Checkmark & 6.74 & 17.33 & 24.95 & 6.16 & 0.25 & ×\\
 \Checkmark & \XSolidBrush &  \bf{0.53} &  44.33 & 0.38 & 0.46& 0.54& ×\\
 \Checkmark & \Checkmark &  0.77 & \bf{2.52} & \bf{0.32}
 &\bf{0.39} & \bf{0.12} & \bf{0.25}
 \\
\bottomrule
\end{tabular}

\begin{tablenotes}
\item[1] Without continuous-time motion model.
\item[2] Using point-to-plane ICP 
in~\cite{zheng2021efficient} for registration.
\item[3] Fail registration in this sequence.
\end{tablenotes}
\end{threeparttable}
\vspace{-0.2in}
\end{table*}

\subsection{Evaluation on Computational Efficiency}

The computational load of our approach is mainly dominated by normal estimation and filter registration. By taking advantage of the effective range image representation, the spatial relationship among points can be used to speed up the millions of latent data associations. To demonstrate the efficiency of our presented method, we evaluate the computational costs at different stages, including data upload from CPU to GPU, normal estimation, filter registration, and map updating. Once the input point cloud is transferred to GPU memory, all the remaining steps are computed on GPU using CUDA. Additionally, we deploy our approach into the embedded devices Jetson AGX, which is wildly used in autonomous driving.
 
The computational time is highly affected by the total number of scattered points within a scan. For fair comparison, we set the output frequency of each LiDAR to 10~Hz. Therefore, the total number of points in single scan is 10K for Livox Mid-40 (`ZJU-M0'), and 24K for Livox Horizon (`ZJU-H0') and Avia (`ZJU-A0'). We make use of all the raw points without downsampling. As depicted in Table~\ref{tab:runtime}, our proposed approach runs around 480 frame per second on the commodity laptop, and 110 frame per second on Jetson AGX. In practice, we do not have to update the map and compute normal for each scan. Instead, we just need update a portion of points by downsampling the raw input data. Thus, our presented scheme can achieve real-time performance for various LiDAR even on a low-end embedded device.

\begin{table}[t]
\vspace{-0.2in}
\footnotesize
\centering
\caption{Evaluation on Computational Time (ms)}

\begin{threeparttable}
\begin{tabular}{@{}lc|cccc|c@{}}
\toprule
    &Dataset & \begin{tabular}[c]{@{}c@{}} Data \\ Upload \end{tabular}
    &\begin{tabular}[c]{@{}c@{}}Normal\\      Estimation\end{tabular}  
    & \begin{tabular}[c]{@{}c@{}}Filter\\      Registration\end{tabular} 
    & \begin{tabular}[c]{@{}c@{}}Map\\      Updating\end{tabular} 
    & \begin{tabular}[c]{@{}c@{}}Total\\      Time\end{tabular}\\

\midrule

\rotatebox[origin=c]{90}{AGX} &\begin{tabular}[c]{@{}l@{}} ZJU-M0\\ ZJU-H0 \\ ZJU-A0 \end{tabular}&
\begin{tabular}[c]{@{}l@{}} 0.36\\ 0.53 \\ 0.57 \end{tabular}&
\begin{tabular}[c]{@{}l@{}} 2.36\\ 4.50 \\ 9.23 \end{tabular}&
\begin{tabular}[c]{@{}l@{}} 4.46\\ 9.33 \\ 7.72 \end{tabular}&
\begin{tabular}[c]{@{}l@{}} 1.87\\ 3.12 \\ 4.75 \end{tabular}&
\begin{tabular}[c]{@{}l@{}} 9.05\\ 17.48 \\ 22.27 \end{tabular} \\
\midrule
\rotatebox[origin=c]{90}{Laptop} &\begin{tabular}[c]{@{}l@{}} ZJU-M0\\ ZJU-H0 \\ ZJU-A0 \end{tabular}& 
\begin{tabular}[c]{@{}l@{}} 0.10\\ 0.19 \\ 0.23 \end{tabular}&
\begin{tabular}[c]{@{}l@{}} 0.51\\ 1.04 \\ 2.19 \end{tabular}&
\begin{tabular}[c]{@{}l@{}} 1.02\\ 2.73 \\ 2.17 \end{tabular}&
\begin{tabular}[c]{@{}l@{}} 0.45\\ 0.77 \\ 1.08 \end{tabular}&
\begin{tabular}[c]{@{}l@{}} 2.08\\ 4.73 \\ 5.67 \end{tabular} \\
\bottomrule
\end{tabular}

\end{threeparttable}
\label{tab:runtime}
\vspace{-0.2in}
\end{table}

\vspace{-0.1in}
\section{CONCLUSION}~\label{sec:conc}
This paper proposed an effective continuous-time odometry approach to LiDAR with small FoV by taking advantage of filter registration. We made use of the continuous-time model to compensate the motion distortions. Moreover, GMM-based filter registration was employed to robustly align sparse point clouds to the noisy map after map updating. Additionally, we employed the efficient range image representation, which not only consumes few memory in map updating but also can be easily implemented in parallel on GPUs. We have conducted the extensive evaluations, whose promising results demonstrated that our proposed approach is very effective.

Currently, our method only takes considerations of LiDAR data, which may lead to the sensor degeneration in some cases. For future work, we will incorporate various sensors into the optimization framework.


\bibliographystyle{IEEEtran}
\bibliography{mybib}

\end{document}